\documentclass[sigconf]{acmart}
\AtBeginDocument{%
  }

\copyrightyear{2026}
\acmYear{2026}
\setcopyright{cc}
\setcctype{by}
\acmConference[HRI '26]{Proceedings of the 21st ACM/IEEE International Conference on Human-Robot Interaction}{March 16--19, 2026}{Edinburgh, Scotland Uk}
\acmBooktitle{Proceedings of the 21st ACM/IEEE International Conference on Human-Robot Interaction (HRI '26), March 16--19, 2026, Edinburgh, Scotland Uk}
\acmPrice{}
\acmDOI{10.1145/3757279.3785589}
\acmISBN{979-8-4007-2128-1/2026/03}

\usepackage{booktabs}
\usepackage{booktabs}
\usepackage{subcaption}
\usepackage{adjustbox}
\usepackage{dblfloatfix}   
\usepackage{placeins}      
\usepackage{algorithm}
\usepackage{amsmath}
\usepackage[noend]{algpseudocode}


\begin{document}

\title[What you reward is what you learn]{What you reward is what you learn: Comparing rewards\\for online speech policy optimization in public HRI}

\author{Sichao Song}
\authornote{Both authors contributed equally to this research.}
\authornotemark[2]
\author{Yuki Okafuji}
\authornotemark[1]
\affiliation{%
  \institution{CyberAgent}
  \city{Tokyo}
  \country{Japan}
}
\additionalaffiliation{%
  \institution{The University of Osaka}
  \city{Osaka}
  \country{Japan}
}
\email{song\_sichao@cyberagent.co.jp}
\email{okafuji\_yuki\_xd@cyberagent.co.jp}

\author{Kaito Ariu}
\affiliation{%
 \institution{CyberAgent}
 \city{Tokyo}
 \country{Japan}}
\email{kaito\_ariu@cyberagent.co.jp}

\author{Amy Koike}
\affiliation{%
 \institution{University of Wisconsin-Madison}
 \city{Wisconsin}
 \country{United States}}
\email{ekoike@wisc.edu}

\renewcommand{\shortauthors}{Sichao and Yuki et al.}

\begin{abstract}

Designing policies that are both efficient and acceptable for conversational service robots in open and diverse environments is non-trivial. Unlike fixed, hand-tuned parameters, online learning can adapt to non-stationary conditions. In this paper, we study how to adapt a social robot’s speech policy in the wild. During a 12-day in-situ deployment with over 1{,}400 public encounters, we cast online policy optimization as a multi-armed bandit problem and use Thompson sampling to select among six actions defined by speech rate (slow/normal/fast) and verbosity (concise/detailed). We compare three complementary binary rewards--\textbf{$R_u$} (user rating), \textbf{$R_c$} (conversation closure), and \textbf{$R_t$} ($\geq$2 turns)--and show that each induces distinct arm distributions and interaction behaviors. We complement the online results with offline evaluations that analyze contextual factors (e.g., crowd level, group size) using video-annotated data. Taken together, we distill ready-to-use design lessons for deploying online optimization of speech policies in real public HRI settings.

\end{abstract}

\ccsdesc[500]{Human-centered computing~Empirical studies in HCI}
\ccsdesc[500]{Human-centered computing~Field studies}
\ccsdesc[300]{Human-centered computing~User models}

\keywords{Human-Robot Interaction (HRI), Conversational Robots, Thompson sampling, Multi-armed Bandits (MAB), Speech Parameter Optimization, Field Experiment, Service Robots}
%

\maketitle

\section{Introduction}

Online learning has increasingly been adopted in Human–Robot Interaction (HRI) to enable personalization and policy adaptation during interaction. For social robots, this extends beyond fixed, hand-tuned scripts: the robot can learn and adapt its action selection in real-time. Such on-the-spot adaptation yields concrete benefits, especially in-the-wild settings: higher task success, smoother turn-taking, better user satisfaction, and less operator burden from constant retuning. In practice, online learning in HRI is realized with multi-armed bandits (MAB) or reinforcement learning (RL), which have repeatedly proven effective for adapting robot behaviors to user preferences across preference learning, persuasive recommendation, tutoring, and socially interactive control \cite{schneider2017exploring,ritschel2019adaptive,ritschel2018drink,cai2021bandit,qureshi2016robot,gao2018robot}. 

Conceptually, MAB differ from full RL in that they maximize cumulative reward over limited trials by balancing exploration and exploitation without an explicit state model. This property---together with extensions for drifting or time-varying preferences---makes MAB attractive for in-the-wild learning where users are heterogeneous and interaction conditions shift over time \cite{kim2024time,mattos2019multi,chen2020fair}. 

Nevertheless, most of the empirical evaluations of MAB applications to social robots remain in closed or semi-controlled settings. Representative studies include dueling bandits for exercise preference adaptation \cite{schneider2017exploring}, an assistive companion that adapts linguistic style from explicit feedback \cite{ritschel2019adaptive}, contextual bandits for personalizing educational chatbots \cite{cai2021bandit}, the Assistive MAB framework formalizing human–robot assistance \cite{chan2019assistive}, and fairness-constrained contextual bandits for multi-user allocation \cite{chen2020fair}. Even recent human-in-the-loop contextual bandits for robot-assisted feeding, while involving real users, are short-horizon and controlled rather than sustained public deployments \cite{banerjee2025ask}. 

By contrast, truly in-the-wild investigations--naturalistic, sustained deployments with walk-up users--remain scarce. An event-scale persuasive drink adviser demonstrated that bandit-driven policy selection can operate amid rapidly changing contexts \cite{ritschel2018drink}. More broadly, the online experimentation literature on ``bandits in the wild" documents practical pitfalls (non-stationarity, delayed rewards, interference) and field-tested strategies that generalize to interactive systems \cite{mattos2019multi}. These signals suggest feasibility while underscoring design debt in reward shaping, exposure control, and contextualization for public HRI.

Taken together, prior work indicates that MAB-based personalization for social robots succeeds in controlled contexts, whereas in-the-wild learning must contend with distributional shift, diverse user behaviors, and imperfect compliance with scripted flows. A central open question is therefore how reward operationalization shapes learned behavior under these conditions: Which observable signals should be rewarded (e.g., task success, engagement, user-reported satisfaction)? How should they be combined? How should exploration be modulated without violating social or fairness constraints \cite{kim2024time,maroto2024personalizing,mattos2019multi,chen2020fair}?

Although some studies focused on reward design compare subjective (explicit) and objective (implicit) feedback and show benefits from combining them \cite{maroto2024personalizing}, prior HRI work has largely explored these ideas in controlled settings. Examples include preference learning with explicit user ratings over repeated encounters~\cite{Baraka2015}; object-fetching that asks users clarifying questions only when needed while leveraging implicit cues~\cite{Whitney2017}; interactive reinforcement learning that integrates task performance (explicit) with task engagement (implicit) to drive real-time personalization~\cite{Tsiakas2018}; socially-aware reinforcement learning that adapts a robot’s linguistic style from user feedback~\cite{Ritschel2019}; and social navigation that jointly plans with both implicit motion-based and explicit multimodal communication~\cite{Che2020}. While these studies demonstrate clear benefits from blending feedback modalities, demonstrations remain predominantly closed-environment, and how different reward definitions steer online learning during long-term, in-the-wild encounters remains underexplored.

This paper targets an in-the-wild commercial facility and systematically compares multiple reward designs within a Thompson-sampling (TS) bandit for the conversational speech policy of a social robot. It is known that users in commercial facilities have diverse attributes, such as age, and that their behavior during interactions varies depending on their psychological state \cite{koike2025drives}, such as motivation. This makes it difficult to create clear rules for selecting actions. TS directly represents uncertainty in reward estimates and uses it to decide when to explore versus exploit. This makes it well-suited to commercial facilities where user attributes are diverse, while keeping the policy free from brittle, hand-engineered selection rules.

We deployed a service robot in a shopping mall for 12 days (over 1{,}400 public encounters), adapting two dimensions--\emph{speech rate} (slow/normal/fast) and \emph{verbosity} (concise/detailed)--while optimizing three binary rewards that capture complementary constructs: $R_u$ (user rating), $R_c$ (conversation closure), and $R_t$ ($\geq$2 turns). Each interaction was logged and video-annotated for social context (e.g., crowd level and group size), enabling post-analysis of how context moderates learned policies. We report (i) TS learning and posterior arm preferences under each reward and (ii) generalized linear models quantifying arm$\times$context interactions. In Thompson sampling, an arm is an action type. In this experiment, each arm refers to a setting of the robot’s speech rate $\times$ verbosity. Based on these findings, we distill design lessons for contextual online optimization in public HRI.

We pursue two objectives in an in-the-wild, public deployment of a conversational service robot:
\begin{itemize}
  \item \textbf{RO1:} Evaluate how alternative reward definitions steer online learning of the speech policy of a social robot. Concretely, we compare $R_u$ (user rating), $R_c$ (conversation closure), and $R_t$ ($\geq$2 turns) within a TS over speech rate (slow\slash normal\slash fast) and verbosity (concise\slash detailed).
  \item \textbf{RO2:} Quantify how social context moderates outcomes and arm effectiveness using video annotations and GLMs; then translate these regularities into actionable future guidance for context-aware online optimization.
\end{itemize}

This paper contributes two complementary advances.
\begin{itemize}
  \item We deploy a social robot in a shopping mall over a 12-day period, demonstrating how TS adapts its speech policy in real time across over 1,400 public encounters. The learned arm preferences diverge by reward design, demonstrating construct-sensitive policy selection in the wild.
  \item We provide video-annotated analysis that quantifies context moderation with GLMs, revealing what factors influence performance outcomes. According to the results, we distill design lessons for context-aware online optimization in public HRI.
\end{itemize}

\section{Methodology}

\subsection{Thompson Sampling (TS)}

Thompson Sampling (TS) is a Bayesian approach to the exploration/exploitation tradeoff, dating back to Thompson’s 1933 proposal of selecting actions in proportion to the posterior probability that they are optimal~\cite{thompson1933likelihood}. In the past decade, TS has received strong theoretical support, including finite time regret analyses that match or approach the best achievable rates~\cite{kaufmann2012thompson}, and it has been widely adopted in online recommendation and advertising~\cite{chapelle2011empirical}. Comprehensive tutorials further situate TS within the bandit literature and show its extensibility to contextual, non-stationary, and constrained settings~\cite{russo2018tutorial}. Key advantages include (i) conceptual and implementation simplicity, (ii) exploration driven by posterior uncertainty without explicit optimism bonuses, and (iii) straightforward accommodation of real world constraints and deployment realities such as delayed rewards and batched updates.

\paragraph{\textbf{Bernoulli rewards with Beta priors.}}
In our setting each interaction yields a binary outcome $r\in\{0,1\}$ (success/failure), so we model the success probability of arm $a$ as $\theta_a\sim\mathrm{Beta}(\alpha_a,\beta_a)$. Because Beta is conjugate to the Bernoulli likelihood, observing reward $r$ after playing arm $a$ updates the posterior via
\[
\alpha_a \leftarrow \alpha_a + r, \qquad
\beta_a \leftarrow \beta_a + (1-r).
\]
At round $t$, TS draws a sample $\tilde{\theta}_a \sim \mathrm{Beta}(\alpha_a,\beta_a)$ for each arm independently and selects $a_t=\arg\max_a \tilde{\theta}_a$. Arms with greater posterior uncertainty are stochastically favored, yielding principled exploration. Common uninformative initializations include $\alpha_a{=}\beta_a{=}1$ (uniform) or a small symmetric prior. In our deployment we preceded learning with a short uniform pre-allocation to ensure minimum exposure (see \S\ref{subsec:exp_design}).
Algorithm~\ref{alg:thompson_bernoulli} shows the standard loop for Bernoulli rewards.

\begin{figure*}[!t]
  \centering
  \captionsetup{font=small,skip=4pt}
  \setlength{\tabcolsep}{0pt}
  \renewcommand{\arraystretch}{1.0}

  \begin{subfigure}[t]{0.45\linewidth}
    \centering
    \setlength{\fboxsep}{2pt}
    \fcolorbox{gray!40}{white}{\adjustbox{max width=0.7\linewidth}{\includegraphics{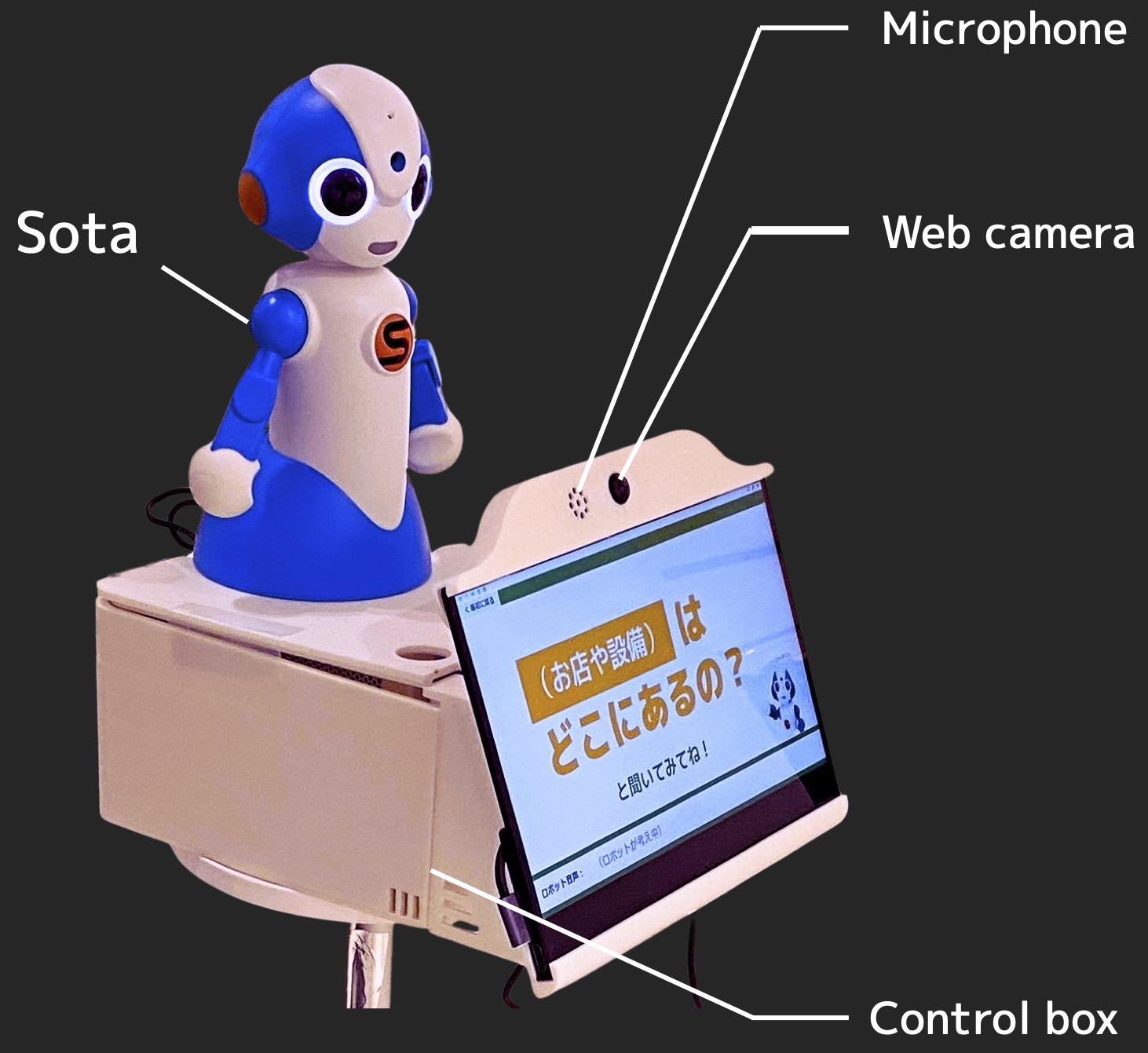}}}
    \caption{Service robot system ``Sota''.}
    \Description{Service robot system.}
    \label{fig:sotasystem}
  \end{subfigure}\hfill
  \begin{subfigure}[t]{0.45\linewidth}
    \centering
    \setlength{\fboxsep}{2pt}
    \fcolorbox{gray!40}{white}{\adjustbox{max width=0.95\linewidth}{\includegraphics{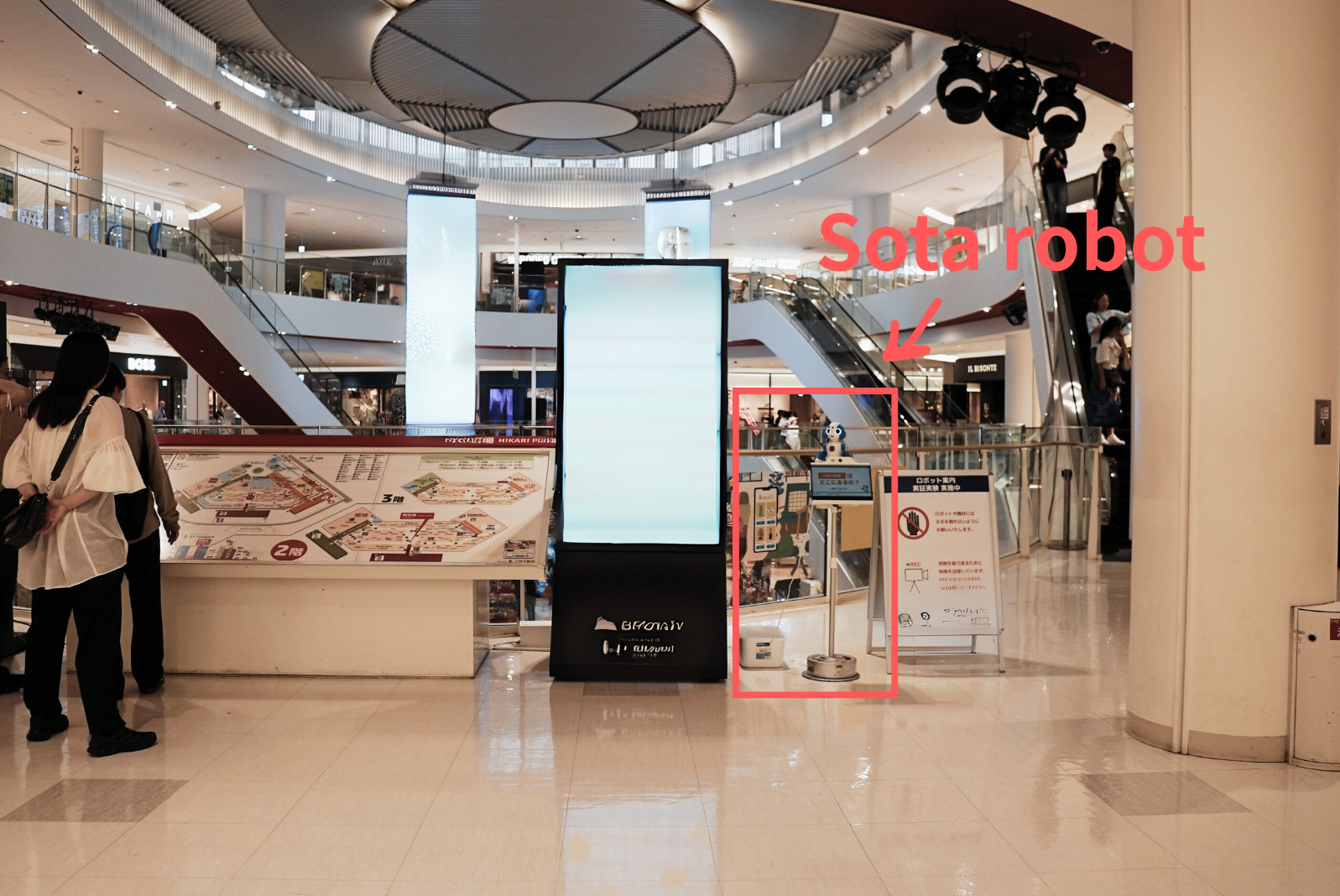}}}
    \caption{Field setting in the shopping mall.}
    \Description{Field setting in the shopping mall.}
    \label{fig:field}
  \end{subfigure}

  \caption{Sota robot designed for route guidance and field setting in the shopping mall. It features an integrated display. The display explicitly communicates the robot’s ability to provide directional assistance, ensuring users are aware of its functionality.}
  \Description{Sota robot designed for route guidance and field setting in the shopping mall. It features an integrated display. The display explicitly communicates the robot’s ability to provide directional assistance, ensuring users are aware of its functionality.}
  \label{fig:sota_field}
\end{figure*}

We used TS to optimize among six arms (slow/normal/fast \texttimes\ \allowbreak concise/detailed). The following engineering choices aligned TS with our deployment setting:
\begin{itemize}
  \item \textbf{Cold start.} We guaranteed minimum exposure by uniformly allocating each arm a small number of initial interactions and seeding $(\alpha_a,\beta_a)$ with those observations.
  \item \textbf{Optional non-stationarity handling.} To discount stale evidence, a simple forgetting factor $\lambda\in(0,1]$ can be applied to sufficient statistics:
  $\alpha_a \leftarrow \lambda \alpha_a + r$, $\beta_a \leftarrow \lambda \beta_a + (1-r)$ (we report results with $\lambda{=}1$ as the default TS).
  \item \textbf{Multiple labels.} We updated the algorithm using three types of binary rewards: a third-party judgement ($R_c,R_t$) based on good/bad judgment, and a user rating on the site ($R_u$). These rewards were delivered immediately after the interaction, and updates were applied upon arrival.
  
\end{itemize}
Together, these practices keep TS simple while addressing non-stationarity, delayed feedback in real-world deployments.

\begin{algorithm}[!t]
  \caption{Thompson Sampling for Bernoulli Rewards}
  \label{alg:thompson_bernoulli}
  \begin{algorithmic}[1]
    \Require Arms $\mathcal{A}$; Beta priors $\{(\alpha_a,\beta_a)\}_{a\in\mathcal{A}}$
    \For{$t = 1,2,\dots$}
      \ForAll{$a \in \mathcal{A}$}
        \State Sample $\tilde{\theta}_a \sim \mathrm{Beta}(\alpha_a,\beta_a)$
      \EndFor
      \State $a_t \gets \arg\max_{a\in\mathcal{A}} \tilde{\theta}_a$
      \State Play $a_t$ and observe $r_t \in \{0,1\}$ (possibly delayed)
      \State \textbf{Update:} $(\alpha_{a_t},\beta_{a_t}) \leftarrow (\alpha_{a_t}+r_t,\; \beta_{a_t}+1-r_t)$
    \EndFor
  \end{algorithmic}
\end{algorithm}

\subsection{Experiment Design}\label{subsec:exp_design}

This study was conducted in collaboration with a shopping mall during normal business operations. The deployment and data handling followed an institutional ethical review and a facility agreement. The experiment was approved by the Research Ethics Committee of The University of Osaka (Reference Number: R1–5–9). This study was conducted on an opt-out basis for unwilling participants who wanted to be removed from the video data.

\paragraph{\textbf{Field and Tasks}}
We conducted the field study at a shopping mall in Japan (details anonymized for double-blind review). The social robot was installed next to the mall's floor map on the second and third floors, as shown in Figure ~\ref{fig:sota_field}. Two conditions were simultaneously operated on the second and third floors, and once one condition was completed, the other condition was implemented. The robot was deployed for 12 days, approximately eight hours per day, until all experimental conditions were achieved. The robot's main task was route guidance, but it was also designed to be able to handle various types of interaction, such as casual conversation.

\paragraph{\textbf{Reward Design}}
Previous research has compared subjective (explicit) and objective (implicit) feedback from users in reward design~\cite{maroto2024personalizing}. However, in real-world environments, user behavior is unique, as social robots are often used as chat partners, despite their intended role as guides. This makes it difficult to define clear objective indicators. Therefore, we firstly prepared four objective indicators to be used in social robot dialogue tasks (conversation closure, overlapping utterances, dialogue conflict, and number of turns) \cite{inoue2025noiserobust} and conducted prior observations to determine whether they were important aspects for online learning of speech policies. After preliminary observation, we decided to conduct the experiment using three types of binary rewards (one subjective and two objective) that reflected different operational goals:

\begin{itemize}
  \item \textbf{$R_u$ (User rating):} success if a post-interaction, single-item satisfaction rating on a 7-point scale was $\geq 6$; failure otherwise.
  \item \textbf{$R_c$ (Conversation closure):} success if the interaction reached an explicit conversational closing routine (farewell and/or thanks); failure otherwise.
  \item \textbf{$R_t$ ($\geq 2$ turns):} success if the dialog contained at least two turns (user-robot exchanges); failure otherwise. An encounter statement is counted from the robot's proximity-triggered greeting, and success requires the minimal sequence R1 (robot) $\rightarrow$ U1 $\rightarrow$ R2 $\rightarrow$ U2; a ``turn'' is a contiguous speaker segment.
\end{itemize}
$R_u$ captures users’ subjective assessments directly, whereas $R_c$ and $R_t$ capture objective aspects of the interaction. For $R_u$, we mapped scores $\geq 6$ to success. According to our previous field studies, scores were positively biased; treating 5 as ``positive'' would result in most interactions being classified as successful. Using the stricter $\geq 6$ threshold yields a more balanced split of positive vs. negative outcomes and provides better discrimination. The remaining two objective rewards were determined by binary criteria in terms of a balanced split of outcomes: whether the conversational closing routine happens and whether the dialogue contained at least two turns.

Our three reward conditions ($R_u, R_c, R_t$) operationalize this explicit--implicit feedback in a public--space deployment, enabling a controlled comparison of how each definition guides the same online learning procedure. In contrast to prior HRI formulations that relied primarily on explicit ratings~\cite{Baraka2015} or combined explicit and implicit signals in lab--style studies~\cite{Whitney2017, Tsiakas2018, Ritschel2019, Che2020}, our study evaluates these alternatives side--by--side under identical hardware, action space, and environment.

\paragraph{\textbf{Data Collection}}
For each reward condition, we used a two-phase schedule: a 30-interaction cold-start phase to seed the posteriors, followed by 450 user interactions with the bandit active. Thus, we obtained 480 interactions per reward in total.

For the cold-start phase, we ran a short uniform pre-allocation: each of the six arms was pulled five times, collecting rewards and updating the per-arm posteriors. We then used the resulting Beta parameters as the bandit’s initial state for Thompson sampling. This ensured a uniform minimum exposure across arms and reduced susceptibility to early mislearning. 

For each interaction we logged: timestamps; the selected arm (speech rate, verbosity); rating of the arm; and bandit variables (e.g., prior/posterior parameters). These logs enable us to reconstruct learning curves and posterior summaries. In addition to logs, we recorded video footage throughout the deployment for annotation purposes.

\begin{figure*}[!t]
  \centering
  \includegraphics[width=0.9\linewidth]{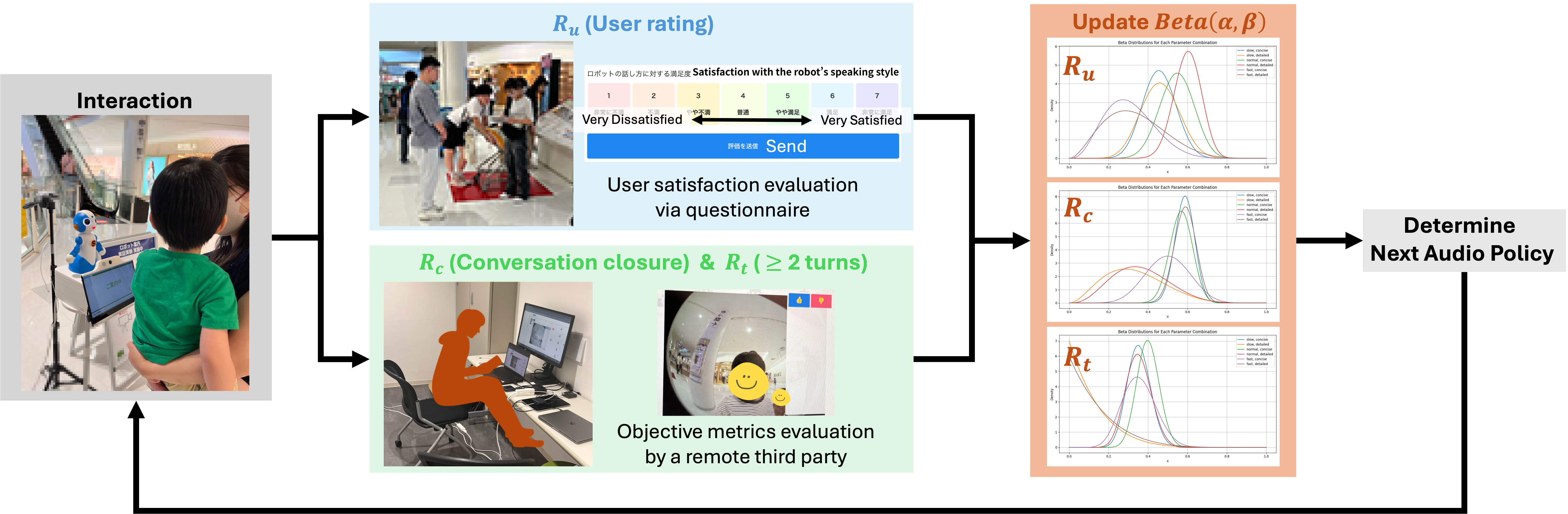}
  \caption{This diagram shows a robot's speech policy change using Thompson sampling (TS). After the interaction ends, three types of reward conditions are given. In the $R_u$ (User rating) condition, the user who interacted with the robot evaluates the robot using a questionnaire, while in the $R_c$ (Conversation closure) and $R_t$ ($\geq$ 2 turns) conditions, a remote third party evaluates the objective metrics of interaction. The obtained evaluation as a reward is sent to the robot system in real time, and the robot's next behavior is determined using TS and immediately reflected.}
  \Description{This diagram shows a robot's speech policy change using Thompson sampling.}
  \label{fig:TSflow}
\end{figure*}

\subsection{Service Robot System}
We used a small humanoid robot, ``Sota'' (Vstone Co., Ltd.), which is \mbox{28\,cm} tall and has a childlike appearance. Each hand has two degrees of freedom (DoF), enabling simple gestures (e.g., a pointing gesture to indicate “that direction”). Sota features three facial LEDs (eyes and mouth) for expressive cues and can rotate its body to realize gaze behaviors. For speech, we used Google Cloud Speech-to-Text (STT) and Text-to-Speech (TTS) APIs. Dialog content generation and dialogue-state management were supported by the OpenAI API (GPT-4 family). A front-mounted display presented short prompts and guidance, including facility maps.

As shown in Fig.~\ref{fig:sotasystem}, a control box beneath Sota houses a mini PC that runs the behavior controller. We implemented a fully autonomous conversational system comprising four basic components: a Recognizer, Dialog Manager, Action Manager, and Modality Manager. Specifically, the Recognizer uses a 180$^\circ$ fisheye camera and PoseNet~\cite{kendall2015posenet} to identify the nearest visitor and infer visit states; the Dialog Manager is a finite-state controller that greets on detection and accepts speech when nearby while grounding LLM-generated content in a curated facility knowledge base and a fixed ``Sota" persona; the Action Manager composes task responses (e.g., route guidance) from dialogue history and the knowledge base; and the Modality Manager triggers about 40 word–gesture mappings—including pointing—to reinforce verbal instructions.

\paragraph{\textbf{Thompson Sampling Integration}}
To enable online optimization of speech policies, we implemented a TS module that selects the arm (speech rate $\times$ verbosity) at the start of each interaction using Thompson sampling for Bernoulli rewards \cite{chapelle2011empirical}. We designed a system in which the robot's behavior is updated in real time through interactions and evaluations. The learning flow using TS is shown in Figure~\ref{fig:TSflow}. The reward signal was derived from two complementary sources: (i) on-site user self-reports for $R_u$ and (ii) third-party success/failure judgments for $R_c$/$R_t$. 

In case (i), users were prompted to complete a survey immediately after the interaction, and their satisfaction with the interaction was rated on a 7-point Likert scale using a survey tablet. A response of 6 or higher was a success, and otherwise was a failure, and the result was immediately sent to the robot system. In case (ii), a monitoring UI streamed camera footage from the robot. The first, second, and fourth authors independently observed interactions in real time and, after the interaction ended, pressed ``Success'' or ``Failure'' according to predefined evaluation criteria. A majority vote instantly sent success and failure labels to the robot system.

The reward results sent to the robot system immediately trigger the TS, which updates the posterior distribution of the arm used in that interaction. The robot's speech policy is then determined based on the updated posterior distribution and immediately reflected in the robot system. The updated behavior is executed the next time the user interacts. This online approach balanced exploration and exploitation, enabling real-time, on-device adaptation of speech policies in the wild.

Regarding detailed speech policy settings, for speech rate that we used Google Cloud TTS rate multipliers slow=$0.80$, normal=$1.20$, and fast=$1.60$; for verbosity, we injected a per-turn instruction into the GPT dialog prompt: concise = ``Respond in short, concise sentences. Focus on the key points and avoid unnecessary elaboration." while detailed = ``Respond in longer, more detailed prose. Include concrete examples and supplementary explanations." Before deployment, the authors piloted all six patterns (arms) on-device and reached consensus that these settings were natural, safe, and sufficiently separable for our field experiment.

\subsection{Evaluation and Analysis}
\paragraph{\textbf{Bandit Performance}}
For each reward condition ($R_u$/$R_c$/$R_t$), we report overall success rates and posterior Beta distributions per arm to reveal learned arm preferences.

Typically, when evaluating a bandit algorithm, the cumulative reward is compared with that of uniform sampling, which performs all actions randomly as a baseline. In our experiment, in addition to running the robot under the three reward conditions, we also ran an experiment under the uniform sampling condition. However, while most of the data for the experiments under the three reward conditions was collected on weekdays, data for the uniform sampling condition was collected mainly on holidays. As a result, we concluded that a fair comparison was difficult due to significant differences in user attributes and behavior patterns. Therefore, in this study, we only compared the speech policies learned under each reward condition as mentioned in ROs, and the comparison with uniform sampling is reported in the Appendix.

\paragraph{\textbf{Video Annotation}}
Apart from analyzing the behaviors acquired in response to different rewards, we also perform a post-hoc analysis to examine how various contextual factors adjust the learned speech policy, for the purpose of discussing future guidance. All 1,400+ interactions were video-recorded and coded to obtain contextual variables. We include the following factors as contextual regressors:
\begin{itemize}
  \item \textbf{Group} (solo/group): whether the focal user approached alone or with companions. Human group presence and characteristics systematically modulate engagement and evaluations toward robots in the wild~\cite{Weiss2010}.
  \item \textbf{Crowd} (present/absent): whether bystanders were visibly near the robot during the interaction. The presence of bystanders alters helping/interaction intentions (bystander effect)~\cite{Liu2025}.
  \item \textbf{Ask-direction} (yes/no): whether the user asked for route direction information. Wayfinding/dialogue needs (e.g., asking for directions) strongly condition dialogue strategies and outcomes in public-space HRI~\cite{Fraune2019}.
  \item \textbf{Motivation} (function/experiment/curiosity/education): the user’s apparent reason for engaging with the robot. The differences in user motivation lead to differences in behavior~\cite{koike2025drives}.
\end{itemize}

In our open-world, multi-party setting, we deliberately exclude demographic covariates (age, gender) from the GLM. Identifying a stable main speaker is unreliable when roles and addressees shift within groups, making per-person demographics methodologically fragile. We therefore model contextual moderators that are observable and theory-motivated--Group, Crowd, Ask-direction, Motivation--rather than individual demographics.

The coding process was primarily carried out by four people, who reviewed and decided on the coding criteria in advance. Because the Group/Crowd/Ask direction are objective indicators of the situation and interaction, there was little variability in the coding criteria between coders. However, because Motivation is internal user information, the coding criteria are more likely to vary depending on the coder. Therefore, for Motivation, around 10\% of the data was checked by multiple coders, resulting in a Cohen's $\kappa$ coefficient of 0.53.

For the binary outcome of reward, generalized linear models (GLMs) were fitted for each reward condition using all of these interaction-coded independent variables:
\begin{equation*}
\begin{aligned}
\text{Outcome} \sim {} & \texttt{arm} + \texttt{crowd} + \texttt{group} + \texttt{ask\_direction} + \texttt{motivation} \\
& {} + \texttt{arm}\times\texttt{crowd} + \texttt{arm}\times\texttt{group}  \\
& {} + \texttt{arm}\times\texttt{ask\_direction} + \texttt{arm}\times\texttt{motivation}.
\end{aligned}
\end{equation*}

Using these analyses, we discuss how the interaction context and user behavior may have influenced the learning outcomes of the speech policy obtained through TS. The results of the GLMs should be important for the future development of in-the-wild online learning frameworks such as MAB.

\begin{table*}[!t]
    \centering
    \captionsetup{font=small,skip=3pt}
    \caption{Descriptive statistics by each reward ($R_u, R_c, R_t$) for the six audio-policy arms [SC = Slow--Concise, SD = Slow--Detailed, NC = Normal--Concise, ND = Normal--Detailed, FC = Fast--Concise, FD = Fast--Detailed]. For each reward definition, both the cold-start sample and the full-deployment sample: trials n / successes / success rate, along with the proportion selected by TS (chosen rate).}
    \label{tab:desc_all_rewards_cs}
    \setlength{\tabcolsep}{3pt}\renewcommand{\arraystretch}{0.92}
    
    \begin{subtable}[t]{0.32\linewidth}
        \centering\scriptsize
        \caption{$R_u$ (user rating)}
        \label{tab:desc_c2_cs}
        \adjustbox{max width=\linewidth}{%
        \begin{tabular}{lcccccc}
            \toprule
            & \multicolumn{2}{c}{Cold start} & \multicolumn{4}{c}{All data} \\
            Arm & $n$ & succ. & $n$ & succ. & succ. rate \% & chos. rate \% \\
            \midrule
            SC & 5 & 3 & 44 & 18  & 40.9 & 9.2 \\
            SD & 5 & 4 & 39 & 19  & 48.7 & 8.1 \\
            NC & 5 & 3 & 98 & 62  & 63.3 & 20.4 \\
            ND & 5 & 4 & 240 & 146 & 60.8 & 50.0 \\
            FC & 5 & 3 & 16 & 6  & 37.5 & 3.3 \\
            FD & 5 & 1 & 43 & 24 & 55.8 & 9.0 \\
            \bottomrule
        \end{tabular}}%
        \end{subtable}\hfill
        \begin{subtable}[t]{0.32\linewidth}
        \centering\scriptsize
        \caption{$R_c$ (conversation closure)}
        \label{tab:desc_c1_cs}
        \adjustbox{max width=\linewidth}{%
        \begin{tabular}{lcccccc}
            \toprule
            & \multicolumn{2}{c}{Cold start} & \multicolumn{4}{c}{All data} \\
            Arm & $n$ & succ. & $n$ & succ. & succ. rate \% & chos. rate \% \\
            \midrule
            SC & 5 & 3 & 96 & 32 & 33.3 & 20.0 \\
            SD & 5 & 0 & 9  &  0 & 0.0 & 1.9 \\
            NC & 5 & 4 & 148& 58 & 39.2 & 30.8 \\
            ND & 5 & 3 & 105& 38 & 36.2 & 21.9 \\
            FC & 5 & 3 & 114& 44 & 38.6 & 23.8 \\
            FD & 5 & 0 & 8  &  0 & 0.0 & 1.7 \\
            \bottomrule
        \end{tabular}}%
    \end{subtable}\hfill
    \begin{subtable}[t]{0.32\linewidth}
        \centering\scriptsize
        \caption{$R_t$ ($\geq$2 turns)}
        \label{tab:desc_c6_cs}
        \adjustbox{max width=\linewidth}{%
        \begin{tabular}{lcccccc}
            \toprule
            & \multicolumn{2}{c}{Cold start} & \multicolumn{4}{c}{All data} \\
            Arm & $n$ & succ. & $n$ & succ. & succ. rate \% & chos. rate \% \\
            \midrule
            SC & 4 & 4 & 198 & 118 & 59.6 & 42.4 \\
            SD & 3 & 0 & 8   & 2   & 25.0 & 1.7 \\
            NC & 3 & 3 & 99  & 55  & 55.6 & 21.2 \\
            ND & 3 & 1 & 18  & 9   & 50.0 & 3.9 \\
            FC & 3 & 2 & 27  & 12  & 44.4 & 5.8 \\
            FD & 3 & 3 & 117 & 70  & 59.8 & 25.1 \\
            \bottomrule
        \end{tabular}}%
    \end{subtable}
\end{table*}

\begin{figure*}[!t]
  \centering
  \captionsetup{font=small,skip=3pt}
  \setlength{\tabcolsep}{2pt}

  \begin{subfigure}[t]{0.32\linewidth}
    \centering
    \adjustbox{max width=\linewidth}{\includegraphics{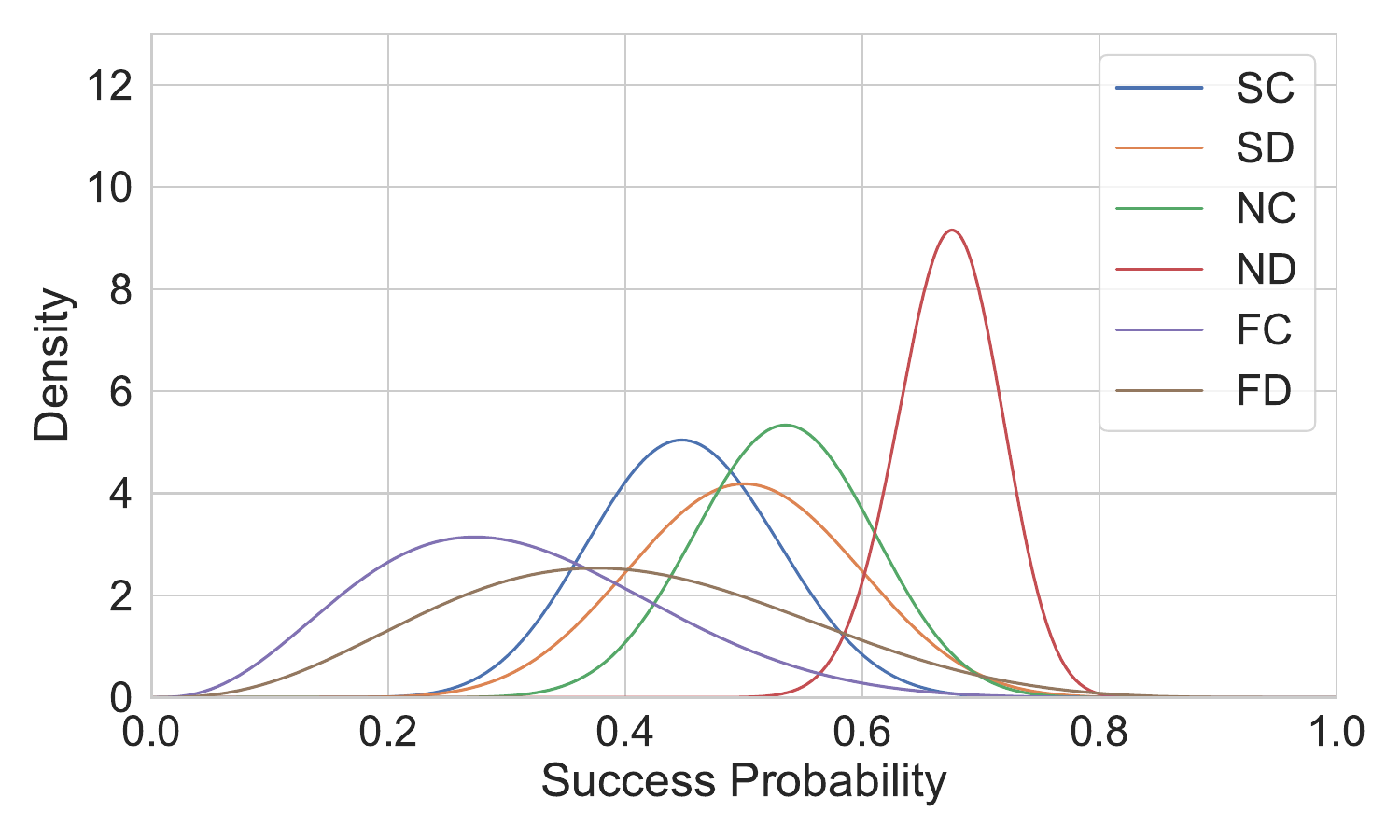}}
    \caption{$R_u$ (user rating) with half-data results}
    \Description{Ru with half-data results.}
    \label{fig:post_c2}
  \end{subfigure}\hfill
  \begin{subfigure}[t]{0.32\linewidth}
    \centering
    \adjustbox{max width=\linewidth}{\includegraphics{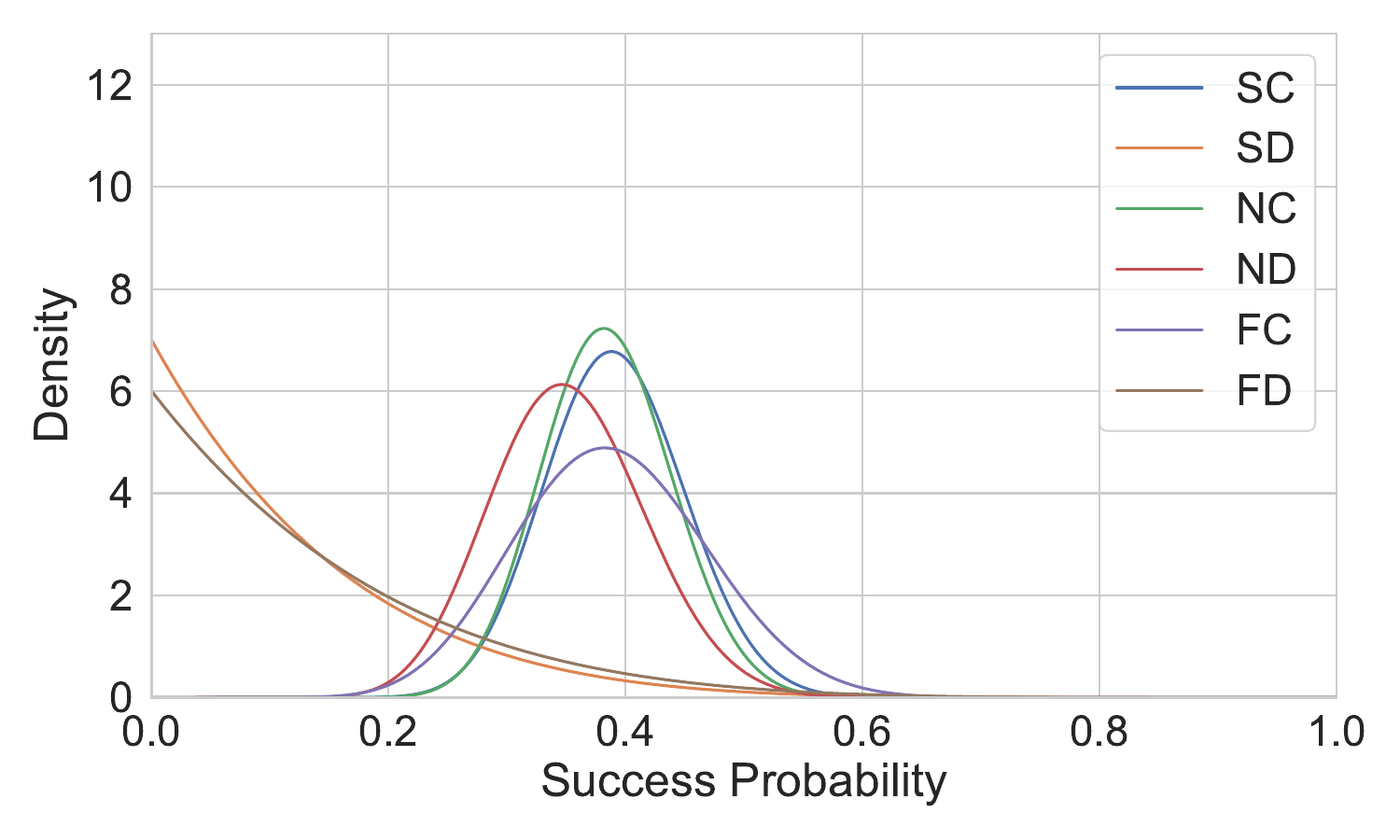}}
    \caption{$R_c$ (conversation closure) with half-data results}
    \Description{Rc with half-data results.}
    \label{fig:post_c1}
  \end{subfigure}\hfill
  \begin{subfigure}[t]{0.32\linewidth}
    \centering
    \adjustbox{max width=\linewidth}{\includegraphics{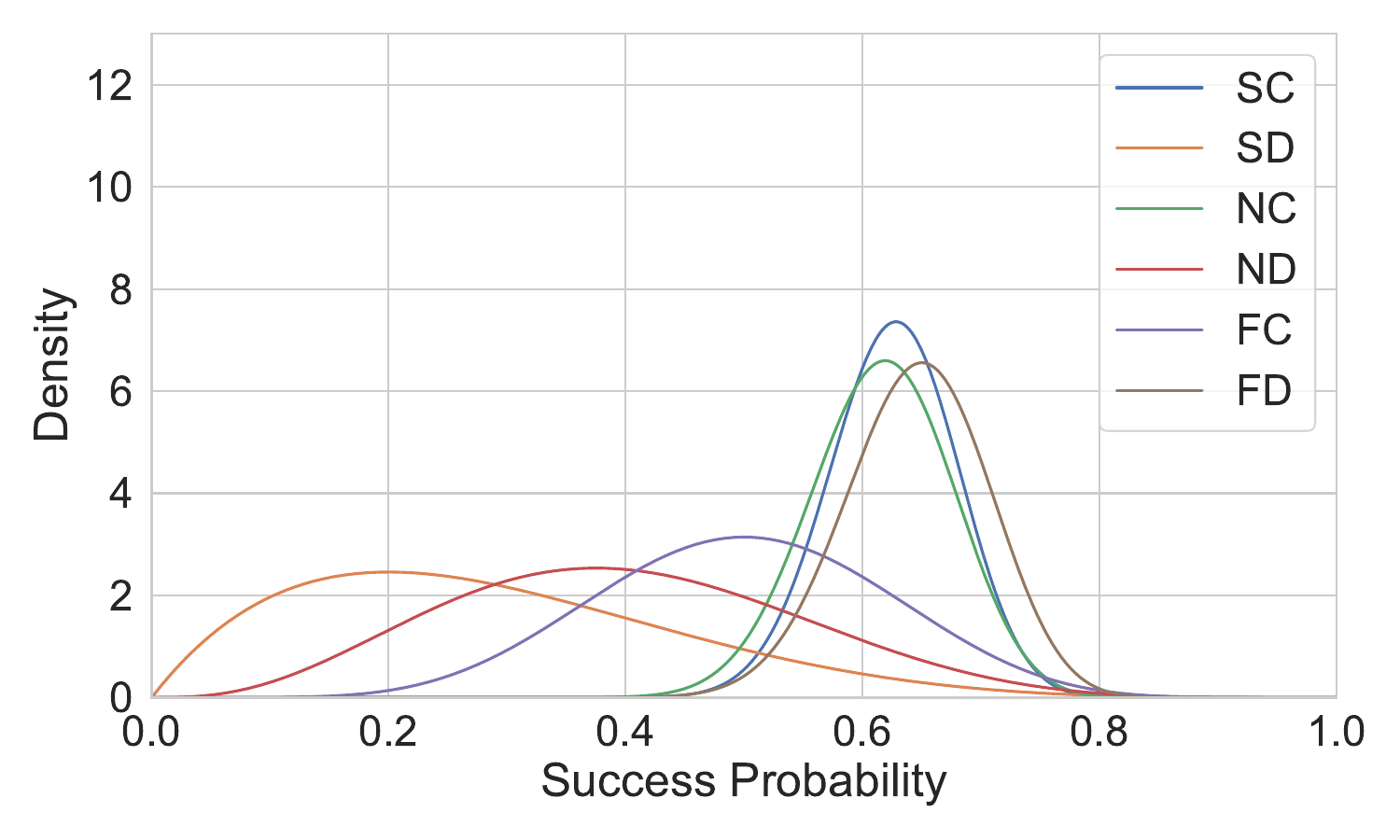}}
    \caption{$R_t$ ($\geq$2 turns) with half-data results}
    \Description{Rt with half-data results.}
    \label{fig:post_c6}
  \end{subfigure}

  \vspace{0.5em}

  \begin{subfigure}[t]{0.32\linewidth}
    \centering
    \adjustbox{max width=\linewidth}{\includegraphics{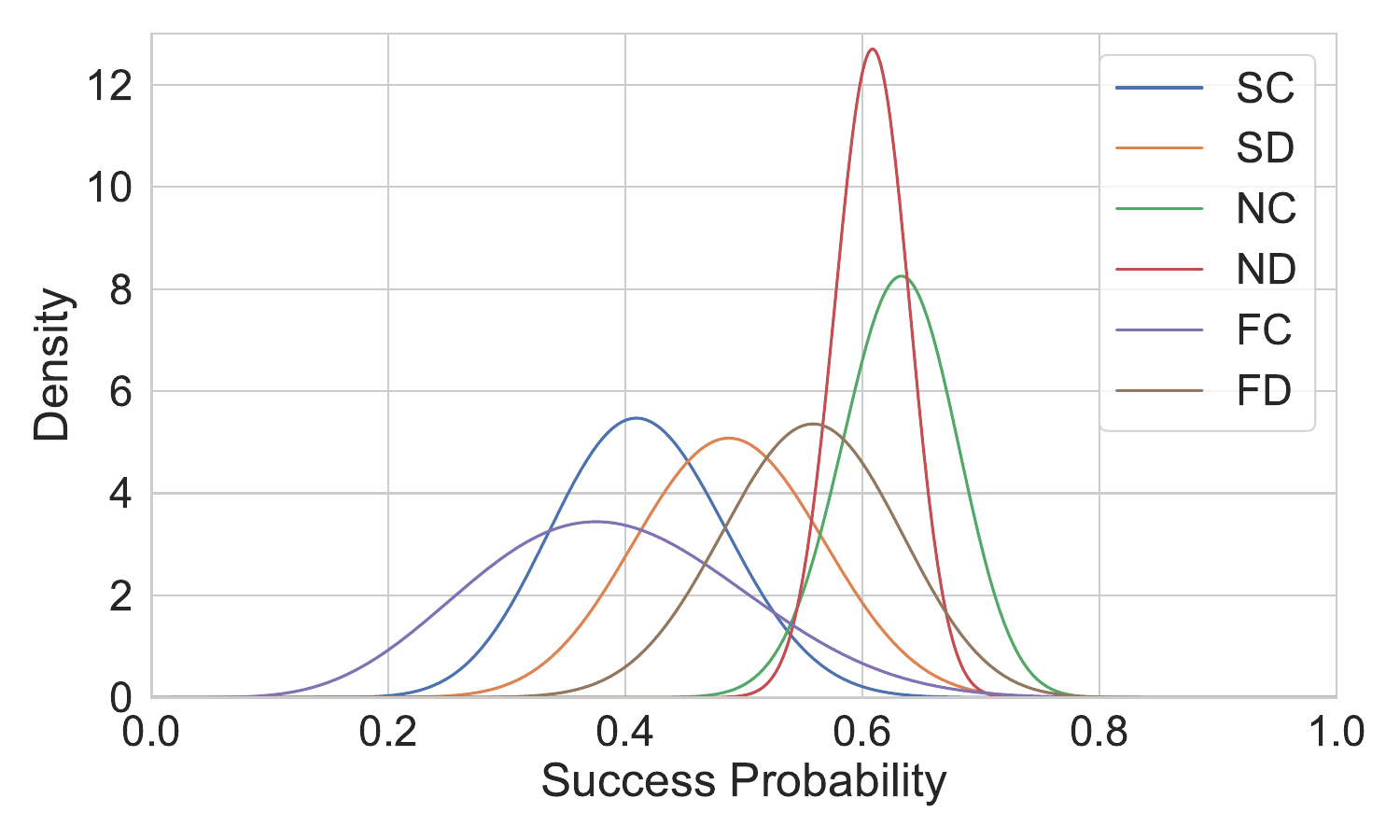}}
    \caption{$R_u$ (user rating) with full-data results}
    \Description{Ru with full-data results.}
    \label{fig:post_c4}
  \end{subfigure}\hfill
  \begin{subfigure}[t]{0.32\linewidth}
    \centering
    \adjustbox{max width=\linewidth}{\includegraphics{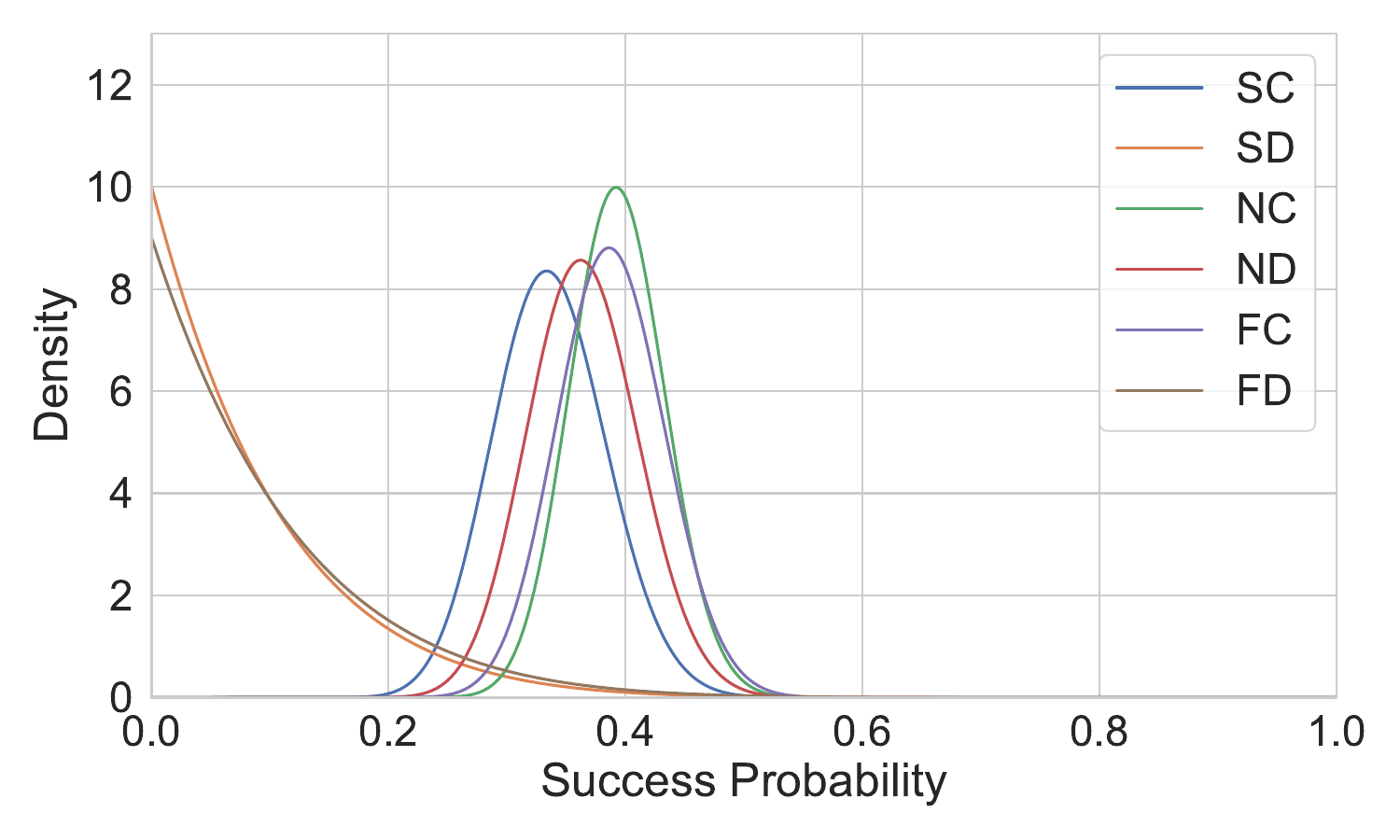}}
    \caption{$R_c$ (conversation closure) with full-data results}
    \Description{Rc with full-data results.}
    \label{fig:post_c5}
  \end{subfigure}\hfill
  \begin{subfigure}[t]{0.32\linewidth}
    \centering
    \adjustbox{max width=\linewidth}{\includegraphics{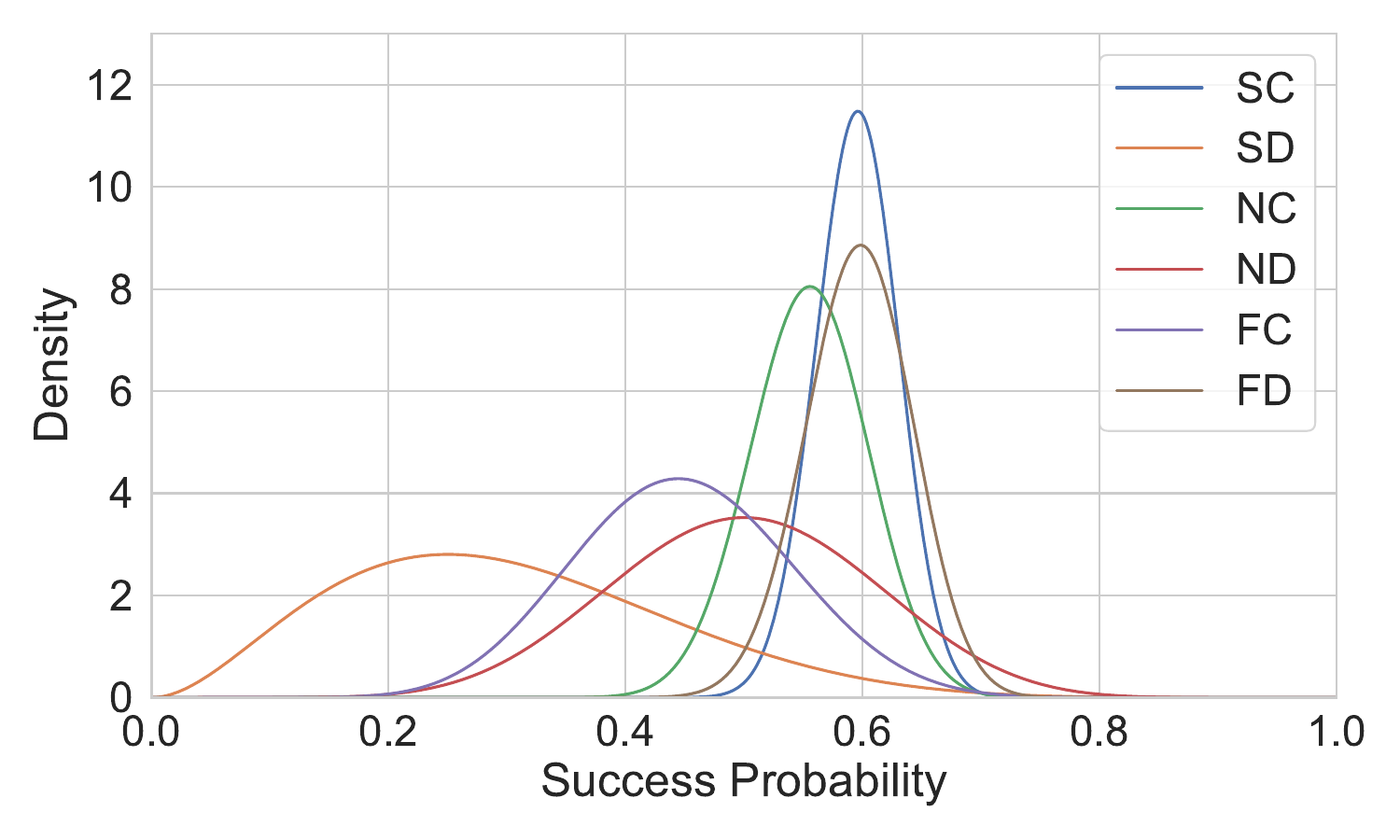}}
    \caption{$R_t$ ($\geq$2 turns) with full-data results}
    \Description{Rt with full-data results.}
    \label{fig:post_c6b}
  \end{subfigure}

  \caption{Posterior Beta densities of per-arm success probability by each reward condition, using half and full of the training data. [SC = Slow--Concise, SD = Slow--Detailed, NC = Normal--Concise, ND = Normal--Detailed, FC = Fast--Concise, FD = Fast--Detailed]}
  \Description{Posterior Beta densities of per-arm success probability by each reward condition, using half and full of the training data.}
  \label{fig:posteriors_all}
\end{figure*}

\section{Results}
\subsection{Bandit Performance}

\paragraph{\textbf{Arm abbreviations (speech rate × verbosity):} }
SC = Slow-Concise, SD = Slow-Detailed, 
NC = Normal-Concise, ND = Normal-Detailed, 
FC = Fast-Concise, FD = Fast-Detailed.

\paragraph{\textbf{Overview}} Table~\ref{tab:desc_all_rewards_cs} reports the descriptive variables (count, success rate, and choice rate) for each reward condition. Cold starts were performed five times in each arm under each condition, except for the $R_t$ condition, where missing data meant that not all arms were performed five times. As can be seen from the All Data values in Table~\ref{tab:desc_all_rewards_cs}, the arms with high success rates had high chosen rates, while the arms with low success rates had low chosen rates, demonstrating the exploration-exploitation characteristic of TS. Thus, although the $R_t$ condition experienced missing cold starts data, it is unlikely to have had a significant impact on the overall data.

Figure~\ref{fig:posteriors_all} shows the posterior beta probabilities, summarizing the arm posterior probabilities learned across half and all data. Showing half the data allows us to see the learning process. A larger expected value of the distribution indicates a higher success rate in each reward condition. The greater the number of selections, the smaller the variance of the distribution, reflecting the results in Table~\ref{tab:desc_all_rewards_cs}.

\paragraph{\textbf{$R_u$ (User rating)}}
A total of 480 data are analyzed. The arm with the highest chosen rate was ND (50.0\%), while the arm with the highest success rate was NC (63.3\%). This can be seen from the learning process in Figure~\ref{fig:posteriors_all}. In the first half of the data, ND has a higher success rate than the other arms and is therefore chosen with a high probability. However, after learning all the data, NC shows the highest success rate. Therefore, NC begins to be chosen more frequently towards the second half of the data, resulting in the highest success rate. Furthermore, among the other arms, the highest chosen and success rates are FD, SD, SC, and FC, in that order, after NC and ND. Compared to the other reward conditions, the variance across all arms is small, indicating that all arms are selected most evenly across all reward conditions.

\paragraph{\textbf{$R_c$ (Conversation closure)}}
A total of 480 data are analyzed. The arm with the highest chosen and success rates is NC (chosen rate: 30.8\%, success rate: 39.2\%). While NC performed best, SC, ND, and FC have comparable chosen and success rates, indicating that learning occurs at a rate roughly equivalent to NC from the early stages of learning. Furthermore, for SD and FD, the success rates are 0.0\% at the cold start, and although they are occasionally chosen exploratorily thereafter, their success rates remain 0.0\% throughout the entire data.

\paragraph{\textbf{$R_t$ ($\geq 2$ turns)}}
A total of 467 data sets were analyzed. The highest chosen rate is SC (42.4\%), and the highest success rate is FD (59.8\%). This is likely due to changes in the success rate of each arm over the learning process, similar to the $R_u$ condition. Following these two arms, the NC arm has the highest chosen and success rates, with the remaining arms showing lower chosen and success rates.

\subsection{GLMs for Modulation}
\setlength{\textfloatsep}{8pt plus 2pt minus 2pt}

\begin{table}[t]
    \centering
    \begingroup
    \scriptsize
    \setlength{\tabcolsep}{4pt}
    \renewcommand{\arraystretch}{1.15}
    \caption{Selected GLM effects across reward definitions (significant predictors only).  Baselines: Arm = \textbf{SC}, Crowd = \textbf{no}, Group = \textbf{group}, Ask-direction = \textbf{no}, Motivation = \textbf{function}. Coefficients are GLM estimates (log-odds). }
    \label{tab:glm_select}
    \begin{tabular}{lccccc}
        \toprule
         & \multicolumn{5}{c}{$R_u$ (user rating)} \\
        \cmidrule(lr){2-6}
        Predictor & $\beta$ & SE & $z$ & $p$ & OR \\
        \midrule
        grp\_single\_T & -1.54 & 0.77 & -1.99 & 0.05* & 0.21 \\
        ND:grp\_single\_T & 1.95 & 0.90 & 2.16 & 0.03* & 7.03 \\
        NC:motivation\_education & 2.76 & 1.41 & 1.96 & 0.05* & 15.76 \\
        ND:motivation\_education & 2.65 & 1.25 & 2.12 & 0.03* & 14.18 \\
        FD:motivation\_education & 2.91 & 1.45 & 2.00 & 0.05* & 18.30 \\
        \midrule
         & \multicolumn{5}{c}{$R_c$ (conversation closure)} \\
        \cmidrule(lr){2-6}
        Predictor & $\beta$ & SE & $z$ & $p$ & OR \\
        \midrule
        NC:grp\_single\_T & 2.68 & 1.31 & 2.04 & 0.04* & 14.59 \\
        \midrule
         & \multicolumn{5}{c}{$R_t$ ($\geq 2$ turns)} \\
        \cmidrule(lr){2-6}
        Predictor & $\beta$ & SE & $z$ & $p$ & OR \\
        \midrule
        ND:crowd\_T & -1.73 & 0.62 & -2.80 & 0.01* & 0.18 \\
        FC:crowd\_T & -1.39 & 0.52 & -2.69 & 0.01* & 0.25 \\
        SD:motivation\_education & 2.33 & 1.16 & 2.01 & 0.04* & 10.26 \\
        FC:motivation\_education & 1.89 & 0.93 & 2.04 & 0.04* & 6.63 \\
        \bottomrule \\
        \multicolumn{6}{r}{\footnotesize * indicates significant difference.}\\
    \end{tabular}
    \endgroup
\end{table}

For each binary outcome according to the reward conditions, we fit binomial-logit GLMs with predictors and arm-context interactions. We report selected statistical summaries for $R_u$/$R_c$/$R_t$ across arms, as shown in Table~\ref{tab:glm_select}; full GLM outputs are provided in the Appendix.

\paragraph{\textbf{Coding and baselines.}}
Arms are treatment-coded with baseline SC (Slow-Concise). 
\texttt{crowd\_T}$=1$ indicates crowd present (baseline: no crowd); 
\texttt{grp\_single\_T}$=1$ indicates single user (baseline: group);
\texttt{ask\_directions\_T}$=1$ indicates asked for directions (baseline: no);
\texttt{motivation} baseline is function. 

\paragraph{\textbf{$R_u$ (user rating).}}
Overall fit: Null deviance = 1184.7 (df = 869) decreased to residual deviance = 1103.2 (df = 828); AIC = 1187.2.
Large standard errors on some arm coefficients indicate partial separation in rare arm-context strata; we focus on stable contextual contrasts and interactions.

\emph{Main effects.}
Single-user interactions yielded lower ratings than group encounters ($p{<}0.05$, OR{=}0.21). There was no overall main effect of crowd, ask-direction, or motivation.

\emph{Interactions.}
In single-user scenes, ND is favored relative to the SC baseline ($p{<}0.05$, OR{=}7.03). For education-motivated visitors, NC, ND, and FD each raise the odds of a high rating ($p{<}0.05$; OR{=}15.76, 14.18, and 18.30).

\paragraph{\textbf{$R_c$ (conversation closure).}}
Overall fit: Null deviance = 592.5 (df = 449) decreased to residual deviance = 524.4 (df = 415); AIC = 594.4. Several interaction terms were singular (NA), indicating sparse arm–context cells.

\emph{Main effects.}
No main effect has a significant difference.

\emph{Interactions.}
A single significant interaction NC$\times$grp\_single\_T ($p{<}0.05$, OR{=}14.59) suggests that in one-to-one encounters, Normal–Concise increased closure odds relative to SC. No other arm context interactions were significant.

\paragraph{\textbf{$R_t$ ($\geq$2 turns).}}
Overall fit: Null deviance = 1218.7 (df = 891) decreased to residual deviance = 1156.3 (df = 852); AIC = 1236.3.
Two interaction terms were singular (NA).

\emph{Main effects.}
No main-effect covariate reached significance.

\emph{Interactions}
In crowd scenes (relative to SC with no crowd), ND and FC are comparatively disadvantaged: ND$\times$crowd ($p{<}.01$, OR{=}0.18) and FC$\times$crowd ($p{<}.01$, OR{=}0.25). For education-motivated users (vs. function-motivated), SD and FC perform better than SC: SD{=}education ($p{<}.05$, OR{=}10.26) and FC$\times$education ($p{<}.05$, OR{=}6.63).

\section{Discussion}

\subsection{Summary of Results}

From our 12-day in-the-wild deployment involving over 1,400 interactions, three key takeaways emerge.

Firstly, the data are not evenly distributed across arms. As posteriors concentrate, the policy favors promising arms and samples dominated ones less; moreover, arms with higher observed success rates are selected more often, consistent with Thompson Sampling’s intended behavior.

Second, arm preference is sensitive to how success is defined; the online learning converged to different speech policies depending on whether success was defined as perceived interaction satisfaction ($R_u$), conversation closure ($R_c$), or conversational persistence ($R_t$). The results indicate that, among well-sampled arms, the best arm is NC (Normal--Concise) and ND (Normal--Detailed) for $R_u$, NC leads for $R_c$, and SC (Slow--Concise) and FD (Fast--Detailed) for $R_t$. 

Third, social context could shape outcomes and interact with speech policies.
\begin{itemize}
  \item Under $R_u$, single-user encounters rate lower than group encounters (main effect). ND especially suits single users. In education-motivated visits, NC/ND/FD tend to be rated higher.
  \item Under $R_c$, reward by conversation closure is broadly stable across context, but single-user encounters specifically favor a normal–concise delivery.
  \item Under $R_t$, in crowded scenes ND and FC underperform. By contrast, education-motivated visits favor SD and FC over other settings.
\end{itemize}

Taken together, (i) the bandit behaved as intended (higher-success arms received more pulls), (ii) the definition of “success” steers which policy emerges as optimal, and (iii) social context moderates these effects. Practically, operators should choose reward definitions aligned with their operational goal (satisfaction vs. closure vs. persistence) and deploy context-aware policies (e.g., for satisfaction, use ND for single users and consider NC/ND/FD in education-motivated cases; to maximize closure with single users, use normal–concise; to sustain conversations, avoid ND/FC in crowds and leverage SD/FC for education-motivated visitors).

\subsection{Speech Policy vs. Reward Design}
Our results demonstrate that the definition of a reward signal fundamentally alters the learned behavior, leading to distinct optimal policies. This underscores that there is no single ``best'' speech policy, but rather a policy that is optimal for a specific operational goal. The observed policy differences are consistent with reports that varying the mix of explicit and implicit user feedback can systematically alter learned behavior~\cite{Baraka2015, Tsiakas2018}. Our contribution is to demonstrate this effect in a public deployment while directly comparing multiple reward definitions for the very same robot and action set.

For the $R_u$ (User Rating) reward, the bandit converged on Normal-Concise (NC) and Normal-Detailed (ND) policies. This suggests that user satisfaction in this public setting is maximized by a normative and predictable interaction style. A normal speech rate is familiar and easy to process, avoiding the potential for frustration from a slow pace or the cognitive load of a fast one. The split between concise and detailed likely reflects differing user preferences for information density, but both fall within a comfortable, non-extreme range. The GLM refines this picture in two ways. First, single-user encounters yield lower ratings overall than group encounters, but ND is particularly effective for single users, consistent with the idea that a more thorough, seemingly personalized response is appreciated when no audience is waiting. Second, education-motivated visits amplify positive evaluations for NC, ND, and FD, suggesting that information-seeking users value either a clear, compact explanation (NC), a fuller step-by-step response (ND), or even fast, information-dense delivery (FD) when they are already motivated to process content.

The $R_c$ (Conversation Closure) reward, which is predicated on reaching a clean end to the interaction, strongly favored the Normal-Concise (NC) policy. This reward operationalizes task efficiency. The NC arm provides information directly and without extraneous detail, enabling users to accomplish their goal and conclude the interaction smoothly. 0.0\% success rate for Slow-Detailed (SD) and Fast-Detailed (FD) arms under this reward is telling; detailed responses likely prolong the interaction unnecessarily or introduce conversational threads that prevent a simple closure, thus failing the reward condition. The GLM sharpens this picture: NC is especially effective in one-to-one encounters.

Finally, the $R_t$ ($\geq2$ turns) reward, a proxy for engagement, yielded a more complex bimodal preference for Slow-Concise (SC) and Fast-Detailed (FD). This suggests two distinct mechanisms for fostering sustained interaction. The SC policy, with its deliberate pacing, may make the robot seem more careful or accessible, potentially prompting users to ask clarifying questions or feel less rushed, thereby extending the dialogue. Conversely, the FD policy may enhance engagement through a different channel: novelty and information density. A fast, detailed response can be more entertaining and provide more conversational hooks for a user to latch onto, sparking curiosity and follow-up questions~\cite{Pang2025}. The GLM results supplement this, showing that ND and FC are comparatively disadvantaged in crowded scenes.

\subsection{Design Lessons for Context-Aware Online Optimization}

In this section, we argue that reward design—rather than parameter tuning alone—primarily shapes which speech policies emerge. The same reasoning extends beyond speech rate and verbosity to a broader action space (e.g., gaze, gesture, display cues), motivating a shift from average-case optimization to context-sensitive control.

Our findings, especially the strong moderation of arm performance by social context, suggest that effective online optimization in HRI must move beyond simple MAB frameworks. While the bandit learned reasonable average policies, the greater opportunity is to adapt policies dynamically to the evolving context of each encounter. Rather than prescribing fixed rules, we outline key considerations for the next generation of context-aware learning systems.

First, the richness of contextual features is paramount. Our results identified crowd level, group size, and user intent as critical variables. However, these are just a starting point. Future systems could benefit from incorporating a much wider array of contextual signals, such as the user's emotional expression~\cite{Spezialetti2020}, or even a memory of past interactions with that individual~\cite{Jens2013}. The challenge and opportunity lie in developing robust, real-time sensing capabilities to capture these nuanced features and represent them in a way that is meaningful for a learning algorithm. This moves the design focus from merely selecting an action to deeply understanding the situation in which the action is taken.

Second, we should explore more sophisticated models for policy learning. While contextual bandits are a natural next step, the dynamic and often unpredictable nature of public HRI may call for even more advanced approaches. For instance, models that can handle non-stationarity—the fact that the best policy might change over the course of a day as mall traffic patterns shift—are essential for long-term deployments~\cite{lattimore2020bandit}. Furthermore, as robots become more capable, their actions will start to influence the subsequent state of the interaction, a condition that traditional bandits do not model. This suggests a future trajectory towards full reinforcement learning (RL), where the robot learns not just an immediate action-reward link but a long-term strategy for interaction~\cite{Kober2014}.

Third, the definition and delivery of the reward signal itself present a design space for exploration. Our study compared three distinct, immediate rewards. However, in long-term interactions, success might be better defined by metrics that unfold over time, such as repeat engagement or successful task completion across multiple encounters. This requires frameworks that can handle delayed or sparse rewards~\cite{lattimore2020bandit, andrychowicz2018}. Moreover, incorporating human feedback more directly into the reward function, for example, by learning from preferences or corrections, could allow for more aligned and personalized robot behavior~\cite{christiano2023}.

This work serves as a stepping stone, demonstrating that context is not just a moderating factor but the very foundation upon which truly adaptive and intelligent social interaction should be built. The goal is not just to find the single best policy, but to create a system that can fluidly generate the right policy at the right time.

\subsection{Limitations}
This study took place at a commercial mall with one robot platform and six pre-defined speech policies (speech rate$\times$verbosity). Findings may differ with other factors, such as type of voice, acoustics, or task settings. We also chose binary rewards for simplicity and sample efficiency. While interpretable, this may compress nuance. Future work should explore alternative thresholds and compare against continuous or composite outcomes (e.g., conversational persistence combined with satisfaction). Furthermore, some GLM coefficients were not estimable because certain arm-context combinations were rarely observed. Longer deployments, if possible, could reduce this sparsity. Finally, our evidence is majorly based on descriptive summaries and GLMs. Stronger claims may benefit from off-policy evaluation and comparisons against established baselines.


\balance
\bibliographystyle{ACM-Reference-Format}
\bibliography{sample-base}

\newpage
\appendix
\section{Supplementary Analyses}
\begin{figure*}[!t]
  \centering
  \includegraphics[width=\textwidth]{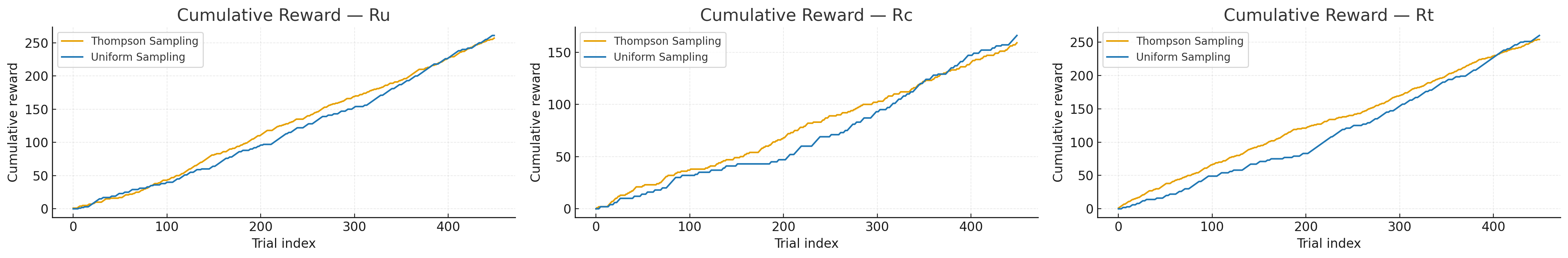}
  \caption{Cumulative reward curves (Thompson Sampling vs.\ Uniform Sampling).}
  \Description{Cumulative reward curves.}
  \label{fig:cumreward-curves}
\end{figure*}

This appendix provides: (1) descriptive statistics comparing Thompson Sampling and Uniform Sampling across all audio-policy arms and rewards ($R_u$, $R_c$, $R_t$) (Table~\ref{tab:desc_ts_uniform}); (2) cumulative reward plots for both allocation schemes by reward (Figure~\ref{fig:cumreward-curves}); and (3) the full GLM results, including non-significant terms and diagnostics (Table~\ref{tab:glm_full_all_one}).

\begin{table*}[!t]
    \centering
    \captionsetup{font=small,skip=3pt}
    \caption{Descriptive statistics comparing Thompson Sampling and Uniform Sampling, stratified by reward ($R_u$, $R_c$, $R_t$), across six audio-policy arms (SC = Slow--Concise, SD = Slow--Detailed, NC = Normal--Concise, ND = Normal--Detailed, FC = Fast--Concise, FD = Fast--Detailed).}
    \label{tab:desc_ts_uniform}
    \setlength{\tabcolsep}{3pt}\renewcommand{\arraystretch}{0.92}
    
    \begin{subtable}[t]{0.32\linewidth}
        \centering\scriptsize
        \caption{$R_u$ (user rating)}
        \label{tab:desc_c2_cs}
        \adjustbox{max width=\linewidth}{%
        \begin{tabular}{lcccccc}
            \toprule
            & \multicolumn{3}{c}{Thompson Sampling} & \multicolumn{3}{c}{Uniform Sampling} \\
            Arm & $n$ & succ. & succ. rate \% & $n$ & succ. & succ. rate \%  \\
            \midrule
            SC & 44 & 18 & 40.9 & 74 & 31 & 41.9 \\
            SD & 39 & 19 & 48.7 & 72 & 29 & 40.3 \\
            NC & 98 & 62 & 63.3 & 78 & 49 & 62.8 \\
            ND & 240 & 146 & 60.8 & 76 & 43 & 56.6 \\
            FC & 16 & 6 & 37.5 & 76 & 56 & 73.7 \\
            FD & 43 & 24 & 55.8 & 74 & 53 & 71.6 \\
            \bottomrule
        \end{tabular}}%
        \end{subtable}\hfill
        \begin{subtable}[t]{0.32\linewidth}
        \centering\scriptsize
        \caption{$R_c$ (conversation closure)}
        \label{tab:desc_c1_cs}
        \adjustbox{max width=\linewidth}{%
        \begin{tabular}{lcccccc}
            \toprule
            & \multicolumn{3}{c}{Thompson Sampling} & \multicolumn{3}{c}{Uniform Sampling} \\
            Arm & $n$ & succ. & succ. rate \% & $n$ & succ. & succ. rate \%  \\
            \midrule
            SC & 96 & 32 & 33.3 & 74 & 20 & 27.0 \\
            SD & 9  &  0 & 0.0 & 72 & 21 & 29.2 \\
            NC & 148& 58 & 39.2 & 78 & 19 & 24.4 \\
            ND & 105& 38 & 36.2 & 76 & 33 & 43.4 \\
            FC & 114& 44 & 38.6 & 76 & 38 & 50.0 \\
            FD & 8  &  0 & 0.0 & 74 & 35 & 47.3 \\
            \bottomrule
        \end{tabular}}%
    \end{subtable}\hfill
    \begin{subtable}[t]{0.32\linewidth}
        \centering\scriptsize
        \caption{$R_t$ ($\geq$2 turns)}
        \label{tab:desc_c6_cs}
        \adjustbox{max width=\linewidth}{%
        \begin{tabular}{lcccccc}
            \toprule
            & \multicolumn{3}{c}{Thompson Sampling} & \multicolumn{3}{c}{Uniform Sampling} \\
            Arm & $n$ & succ. & succ. rate \% & $n$ & succ. & succ. rate \%  \\
            \midrule
            SC & 198 & 118 & 59.6 & 74 & 30 & 40.5 \\
            SD & 8   & 2   & 25.0 & 72 & 37 & 51.4 \\
            NC & 99  & 55  & 55.6 & 78 & 37 & 47.4 \\
            ND & 18  & 9   & 50.0 & 76 & 49 & 64.5 \\
            FC & 27  & 12  & 44.4 & 76 & 51 & 67.1 \\
            FD & 117 & 70  & 59.8 & 74 & 56 & 75.7 \\
            \bottomrule
        \end{tabular}}%
    \end{subtable}
\end{table*}

\begin{table*}[t]
    \centering
    \scriptsize
    \begingroup
    \setlength{\tabcolsep}{4pt}
    \renewcommand{\arraystretch}{1.15}
    \caption{GLM full results across rewards (revised). Baselines: Arm=SC, crowd=no, group=group, ask=no, motivation=function.}
    \label{tab:glm_full_all_one}
    \begin{tabular}{l *{15}{c}}
        \toprule
         & \multicolumn{5}{c}{$R_u$ (user rating)} & \multicolumn{5}{c}{$R_c$ (conversation closure)} & \multicolumn{5}{c}{$R_t$ ($\geq 2$ turns)} \\
        \cmidrule(lr){2-6} \cmidrule(lr){7-11} \cmidrule(lr){12-16}
        Predictor
        & $\beta$ & SE & $z$ & $p$ & OR
        & $\beta$ & SE & $z$ & $p$ & OR
        & $\beta$ & SE & $z$ & $p$ & OR \\
        \midrule
        (Intercept)
        & -14.74 & 1455.40 & -0.01 & 0.99 & 0.00
        & -0.72 & 0.96 & -0.76 & 0.45 & 0.48
        & -0.52 & 0.73 & -0.72 & 0.47 & 0.59 \\
        SD
        & -16.60 & 2520.82 & -0.01 & 0.99 & 0.00
        & -0.99 & 1.14 & -0.87 & 0.38 & 0.37
        & -0.71 & 0.65 & -1.09 & 0.27 & 0.49 \\
        NC
        & 16.33 & 1455.40 & 0.01 & 0.99 & —
        & -15.15 & 1741.84 & -0.01 & 0.99 & 0.00
        & 0.52 & 1.20 & 0.43 & 0.66 & 1.68 \\
        ND
        & 14.49 & 1455.40 & 0.01 & 0.99 & —
        & 18.75 & 2797.44 & 0.01 & 0.99 & —
        & 1.92 & 1.53 & 1.26 & 0.21 & 6.85 \\
        FC
        & 14.32 & 1455.40 & 0.01 & 0.99 & —
        & -17.62 & 1940.86 & -0.01 & 0.99 & 0.00
        & 0.72 & 1.33 & 0.54 & 0.59 & 2.05 \\
        FD
        & 15.96 & 1455.40 & 0.01 & 0.99 & —
        & 0.58 & 0.97 & 0.59 & 0.55 & 1.78
        & 0.48 & 0.94 & 0.51 & 0.61 & 1.61 \\
        crowd\_T
        & 0.57 & 0.61 & 0.92 & 0.36 & 1.76
        & -0.28 & 0.90 & -0.31 & 0.76 & 0.76
        & 0.35 & 0.26 & 1.33 & 0.18 & 1.42 \\
        grp\_single\_T
        & \textbf{-1.54} & \textbf{0.77} & \textbf{-1.99} & \textbf{0.05*} & \textbf{0.21}
        & -0.62 & 0.91 & -0.68 & 0.50 & 0.54
        & -0.06 & 0.39 & -0.16 & 0.87 & 0.94 \\
        ask\_directions\_T
        & 14.87 & 1455.40 & 0.01 & 0.99 & —
        & -0.47 & 0.72 & -0.65 & 0.52 & 0.63
        & 0.59 & 0.69 & 0.85 & 0.40 & 1.80 \\
        experiment
        & -0.40 & 0.51 & -0.78 & 0.44 & 0.67
        & 0.55 & 0.69 & 0.79 & 0.43 & 1.73
        & -0.17 & 0.29 & -0.61 & 0.54 & 0.84 \\
        curiosity
        & -0.83 & 1.31 & -0.63 & 0.53 & 0.44
        & -0.75 & 1.34 & -0.56 & 0.58 & 0.47
        & 0.41 & 0.71 & 0.58 & 0.56 & 1.51 \\
        education
        & -2.11 & 1.19 & -1.77 & 0.08 & 0.12
        & -16.37 & 1978.09 & -0.01 & 0.99 & 0.00
        & -0.80 & 0.48 & -1.67 & 0.10 & 0.45 \\
        \midrule
        SD:crowd\_T
        & 0.05 & 0.82 & 0.07 & 0.95 & 1.05
        & 0.19 & 1.27 & 0.15 & 0.88 & 1.20
        & -1.40 & 0.80 & -1.75 & 0.08 & 0.25 \\
        NC:crowd\_T
        & -0.23 & 0.72 & -0.32 & 0.75 & 0.80
        & -15.93 & 1233.21 & -0.01 & 0.99 & 0.00
        & -0.42 & 0.43 & -1.00 & 0.32 & 0.65 \\
        ND:crowd\_T
        & -0.79 & 0.67 & -1.18 & 0.24 & 0.46
        & -0.46 & 1.25 & -0.37 & 0.72 & 0.63
        & \textbf{-1.73} & \textbf{0.62} & \textbf{-2.80} & \textbf{0.01*} & \textbf{0.18} \\
        FC:crowd\_T
        & 0.18 & 0.82 & 0.22 & 0.83 & 1.19
        & -0.36 & 1.03 & -0.35 & 0.73 & 0.70
        & \textbf{-1.39} & \textbf{0.52} & \textbf{-2.69} & \textbf{0.01*} & \textbf{0.25} \\
        FD:crowd\_T
        & -0.62 & 0.75 & -0.82 & 0.41 & 0.54
        & 0.79 & 1.03 & 0.77 & 0.44 & 2.20
        & -0.30 & 0.42 & -0.71 & 0.48 & 0.74 \\
        SD:grp\_single\_T
        & 0.16 & 1.15 & 0.14 & 0.89 & 1.17
        & 1.15 & 1.40 & 0.82 & 0.41 & 3.15
        & -0.62 & 1.01 & -0.61 & 0.54 & 0.54 \\
        NC:grp\_single\_T
        & 0.98 & 0.93 & 1.05 & 0.29 & 2.67
        & \textbf{2.68} & \textbf{1.31} & \textbf{2.04} & \textbf{0.04*} & \textbf{14.59}
        & -0.05 & 0.69 & -0.07 & 0.94 & 0.95 \\
        ND:grp\_single\_T
        & \textbf{1.95} & \textbf{0.90} & \textbf{2.16} & \textbf{0.03*} & \textbf{7.03}
        & 0.51 & 1.33 & 0.39 & 0.70 & 1.67
        & 0.15 & 1.02 & 0.15 & 0.88 & 1.16 \\
        FC:grp\_single\_T
        & 1.02 & 1.10 & 0.93 & 0.36 & 2.77
        & 0.68 & 1.20 & 0.57 & 0.57 & 1.98
        & -0.04 & 0.77 & -0.05 & 0.96 & 0.96 \\
        FD:grp\_single\_T
        & 0.71 & 1.70 & 0.42 & 0.68 & 2.04
        & --- & --- & --- & --- & ---
        & -0.80 & 0.69 & -1.16 & 0.25 & 0.45 \\
        SD:ask\_directions\_T
        & 16.88 & 2520.82 & 0.01 & 1.00 & —
        & --- & --- & --- & --- & ---
        & --- & --- & --- & --- & --- \\
        NC:ask\_directions\_T
        & -16.24 & 1455.40 & -0.01 & 0.99 & 0.00
        & 14.32 & 1741.84 & 0.01 & 0.99 & —
        & -0.57 & 1.13 & -0.51 & 0.61 & 0.56 \\
        ND:ask\_directions\_T
        & -14.47 & 1455.40 & -0.01 & 0.99 & 0.00
        & -17.59 & 2797.44 & -0.01 & 1.00 & 0.00
        & -0.97 & 1.47 & -0.66 & 0.51 & 0.38 \\
        FC:ask\_directions\_T
        & -13.83 & 1455.40 & -0.01 & 0.99 & 0.00
        & 18.87 & 1940.86 & 0.01 & 0.99 & —
        & 0.24 & 1.27 & 0.19 & 0.85 & 1.28 \\
        FD:ask\_directions\_T
        & -15.24 & 1455.40 & -0.01 & 0.99 & 0.00
        & --- & --- & --- & --- & ---
        & 0.03 & 0.87 & 0.04 & 0.97 & 1.03 \\
        SD:experiment
        & -0.59 & 0.74 & -0.80 & 0.43 & 0.55
        & 0.93 & 1.18 & 0.78 & 0.43 & 2.53
        & 1.31 & 0.70 & 1.87 & 0.06 & 3.70 \\
        NC:experiment
        & 0.71 & 0.63 & 1.12 & 0.26 & 2.03
        & 0.29 & 1.09 & 0.26 & 0.79 & 1.33
        & 0.16 & 0.47 & 0.33 & 0.74 & 1.17 \\
        ND:experiment
        & 0.65 & 0.58 & 1.12 & 0.26 & 1.92
        & -1.01 & 0.94 & -1.07 & 0.29 & 0.36
        & -0.15 & 0.65 & -0.22 & 0.82 & 0.86 \\
        FC:experiment
        & 0.83 & 0.79 & 1.05 & 0.29 & 2.29
        & -0.15 & 0.89 & -0.17 & 0.86 & 0.86
        & -0.28 & 0.55 & -0.51 & 0.61 & 0.75 \\
        FD:experiment
        & 0.09 & 0.78 & 0.12 & 0.90 & 1.10
        & -0.36 & 1.02 & -0.36 & 0.72 & 0.70
        & 0.41 & 0.45 & 0.92 & 0.36 & 1.51 \\
        SD:curiosity
        & 15.98 & 1455.40 & 0.01 & 0.99 & —
        & --- & --- & --- & --- & ---
        & --- & --- & --- & --- & --- \\
        NC:curiosity
        & 1.39 & 1.84 & 0.75 & 0.45 & 4.01
        & --- & --- & --- & --- & ---
        & -0.08 & 1.16 & -0.07 & 0.94 & 0.92 \\
        ND:curiosity
        & 2.26 & 1.65 & 1.37 & 0.17 & 9.54
        & -17.97 & 2797.44 & -0.01 & 0.99 & 0.00
        & -2.51 & 1.95 & -1.28 & 0.20 & 0.08 \\
        FC:curiosity
        & -1.10 & 1.82 & -0.61 & 0.54 & 0.33
        & --- & --- & --- & --- & ---
        & 14.16 & 362.66 & 0.04 & 0.97 & — \\
        FD:curiosity
        & -0.10 & 1.73 & -0.06 & 0.96 & 0.91
        & --- & --- & --- & --- & ---
        & 1.49 & 1.41 & 1.06 & 0.29 & 4.44 \\
        SD:education
        & 17.26 & 650.87 & 0.03 & 0.98 & —
        & 18.15 & 1978.09 & 0.01 & 0.99 & —
        & \textbf{2.33} & \textbf{1.16} & \textbf{2.01} & \textbf{0.04*} & \textbf{10.26} \\
        NC:education
        & \textbf{2.76} & \textbf{1.41} & \textbf{1.96} & \textbf{0.05*} & \textbf{15.76}
        & 35.95 & 3426.15 & 0.01 & 0.99 & — 
        & 1.35 & 0.69 & 1.96 & 0.05 & 3.88 \\
        ND:education
        & \textbf{2.65} & \textbf{1.25} & \textbf{2.12} & \textbf{0.03*} & \textbf{14.18}
        & 17.50 & 1978.09 & 0.01 & 0.99 & — 
        & 1.21 & 0.99 & 1.22 & 0.22 & 3.35 \\
        FC:education
        & 1.42 & 1.41 & 1.01 & 0.31 & 4.14
        & 17.49 & 1978.09 & 0.01 & 0.99 & —
        & \textbf{1.89} & \textbf{0.93} & \textbf{2.04} & \textbf{0.04*} & \textbf{6.63} \\
        FD:education
        & \textbf{2.91} & \textbf{1.45} & \textbf{2.00} & \textbf{0.05*} & \textbf{18.30}
        & 17.16 & 1978.09 & 0.01 & 0.99 & —
        & 1.26 & 0.71 & 1.78 & 0.07 & 3.54 \\
        \bottomrule \\
        \multicolumn{16}{r}{\footnotesize * indicates $p<.05$. \; OR omitted (—) when $|\beta|\gtrsim 8$ due to ill-conditioned estimates.}\\
    \end{tabular}
    \endgroup
\end{table*}

\end{document}